\documentclass[11pt,a4paper]{article}
\usepackage[hyperref]{emnlp2018}
\usepackage{times}
\usepackage{latexsym}
\usepackage{graphicx}
\usepackage{url}
\usepackage{flushend}
\aclfinalcopy 

\usepackage{color}

\title{(Self-Attentive) Autoencoder-based Universal Language Representation \\for Machine Translation}

\author{Carlos Escolano, Marta R. Costa-juss\`a and Jos\'e A. R. Fonollosa \\
  TALP Research Center, Universitat Polit\`ecnica de Catalunya, Barcelona \\
  {\tt \{carlos.escolano,marta.ruiz,jose.fonollosa\}@upc.edu} }

\date{}

\begin{document}
\maketitle
\begin{abstract}
Universal language representation is the holy grail in machine translation (MT). Thanks to the new neural MT approach, it seems that there are good perspectives towards this goal. In this paper, we propose a new architecture based on combining variational autoencoders with encoder-decoders, and introducing an interlingual loss as an additional training objective. 
By adding and forcing this interlingual loss, we are able to train multiple encoders and decoders for each language, sharing a common universal representation. Since the final objective of this universal representation is producing close results for similar input sentences (in any language), we propose to evaluate it by encoding the same sentence in two different languages, decoding both latent representations into the same language and comparing both outputs. Preliminary results on the WMT 2017 Turkish/English task shows that the proposed architecture is capable of learning a universal language representation and simultaneously training both translation directions with state-of-the-art results. 
\end{abstract}

\section{Introduction}

Neural Machine Translation (NMT) \cite{cho:2014}
has arisen as a completely new paradigm for MT outperforming previous statistical approaches \cite{koehn:2003} in most of the tasks. One clear exception are low-resourced tasks \cite{koehn:2017}, where SMT still can outperform NMT. 

Among others, one clear advantage of NMT is that it opens news challenges in MT like multimodal MT \cite{elliott:2017:WMT} or unsupervised MT \cite{artetxe:2017}.
 NMT is progressing fast and it has high expectations, among which there is the finding of a universal language that allows to train single encoders and decoders for each language reducing the number of translation systems from a quadratic dependency on languages to linear. As we will show in section \ref{section2}, there are different approaches that have used the idea of universal language in NMT. However, recent research in this topic has been mainly on evaluating if the NMT architecture of encoder-decoder with recurrent neural networks (RNNs), with or without attention mechanisms, is able to reach a universal language while training multiple languages \cite{johnson:2016,schwenk:2017}. In other words, these approaches have not explicitly designed an architecture with the objective of reaching a universal language representation. 

Differently, in this paper, we specifically pursue towards training a universal language by proposing an architecture combining variational autoencoders and encoder-decoders based on self-attention mechanisms \cite{vaswani2017attention}. Also, in the optimisation process, we are adding a loss term, which is the correlation between intermediate representations from different languages. Like this, we are forcing the system to learn the universal language while training multiple translation systems. Results on the WMT 2017\footnote{http://www.statmt.org/wmt17/} Turkish-English show that our architecture can successfully train the universal language while achieving state-of-the-art translation quality for all translation directions. 

\section{Related Work}
\label{section2}

Classical interlingua approaches \cite{AlAnsary2011InterlinguabasedMT} aim at finding a universal language that involves a conceptual understanding of all languages over the world. This has been the case of Esperanto \cite{Harlow:2013} or Universal Networking Language \cite{unl} and many others. Very differently, in this work, we are focusing on training a universal language representation with deep learning techniques. The objective is to train an intermediate representation that allows to use independent encoders and decoders for each language. 
Differently, from the classical approach, there is no requirement of semantics for this intermediate representation. Following a similar objective or methodology, most related works are the following ones.

\paragraph{Shared Encoders/Decoders.} Johnson et al. (2016) \nocite{johnson:2016} feed a single encoder and decoder with multiple input and output languages. With this approach, authors show that zero-shot learning is possible. Authors show by means of visualizing similar sentences in different languages that there is some hint that these appear somehow close in the common representation.

\paragraph{Dedicated Encoder/Decoder.} These approaches vary from many encoders to one decoder (many-to-one) \cite{zoph:2016}, one encoder to many decoders (one-to-many) \cite{dong:2016} and, finally, one encoder to one decoder (one-to-one), which we are focusing on because they are closest to our approach. Firat et al. 2016 \nocite{firat:2017} propose to extend the classical recurrent NMT bilingual architecture \cite{bahdanau:2015} to multilingual by designing a single encoder and decoder for each language with a shared attention-based mechanism. Schwenk et al. (2017) \nocite{schwenk:2017} and Espana-Bonet et al. (2017) \nocite{espana:2017} evaluate how a recurrent NMT architecture without attention is able to generate a common representation between languages. Authors use the inner product or cosine distance to evaluate the distance between sentence representations. Recently, Lu et al., (2018) \nocite{lu:2017} train single encoders and decoders for each language generating interlingual embeddings which are agnostic to the input and output languages. 

\paragraph{Other related architectures.} While unsupervised MT \cite{lample:2017,artetxe:2017} is not directly pursuing a universal language representation, but it is somehow related with our approach. Artetxe et al. (2017) and Lample et al. (2017) 
propose a translation system that is able to translate trained only on monolingual corpus. The architecture is basically a shared encoder with pre-trained embeddings and two decoders (one of them includes an autoencoder). On the other hand, our work is also related to recent works on sentence representations \cite{conneau:2017,conneau:2018,eriguchi:2018}. However, the main difference is that these works aim at extending representations to other natural language processing tasks, while we are aiming at finding the most suitable representation to make interlingual machine translation feasible. It is left for further work the evaluation and adaptation of this intermediate representation to multiple tasks.

\section{Contribution}

While approaches mentioned in previous section consider the idea of a universal language representation with NMT, they do not really add it in the core training of their architectures. In our architecture, we are adding, as part of the loss function, the correlation between the universal representations. Additionally, we are for the first time proposing a universal language representation within an architecture including self attention mechanisms. 

Another contribution is that we are proposing novel measures to evaluate the universal language: first, in training time, when using the correlation to compare two universal representations, and second, in inference, when using BLEU to compare decoding outputs of two universal representations of the same input sentences coming from two different languages.

\section{Background}

In this section, we report three techniques that are used for the development of the proposed architecture in this paper: variational autoencoders \cite{rumelhart1985learning}, decomposed vector quantization \cite{oord:2017} and correlated nets \cite{chandar:2015}.

\subsection{Variational Autoencoders}

Autoencoders consist in a generative model that is able to generate its own input. This is useful to train an intermediate representation, which can be later employed as feature for another task or even as a dimensionality reduction technique. This is the case of traditional autoencoders that learn to produce an intermediate representation for an existing example. Variational autoencoders \cite{rumelhart1985learning,kingma:2013,zhang-EtAl:2016:EMNLP20162} present a different approach in which the objective is to learn the parameters of a probability distribution that characterizes the intermediate representation. This allows to sample new synthetic instances from the distribution and generate them using the decoder part of the network.

\subsection{Decomposed Vector Quantization}

One of the strategies to create variational autoencoders is vector quantization \cite{oord:2017}. This consists in the addition of a table of dimension $K \cdot D$ where $K$ is the number of possible representations and $D$ the dimension or set of dimensions of each of the representations. The closest vector to the output of the encoder of the network is fed to the decoder as a discrete latent representation to generate the output of the network. 

As proposed in \cite{kaiser2018fast} this approach may produce a vector quantization in which only a small part of the vectors is employed. To solve this, \textit{decomposed vector quantization} uses a set of $n$ tables in which each table is used to represent a portion of the representation that would be later concatenated and fed to the decoder. This approach presents the advantage that by using $n$ $K \cdot \frac{D}{n}$ tables and optimizing the same number of parameters, $K^n$ possible vectors of dimension $D$ can be generated.

\subsection{Correlated Nets}
Chandar et al. (2015) focus on the objective of common representation learning. This work is inspired by the classical Canonical Correlation Analysis \cite{hotelling:1936} and it proposes to use an autoencoder that uses the correlation information to learn the intermediate representation. In this paper, we use this correlation information to measure the distance among intermediate representations following the expression: 

\begin{equation}
\tiny
c(h(X),h(Y)) = \frac{\sum^{n}_{i=1}(h(x_{i} - \overline{h(X)}))(h(y_{i} - \overline{h(Y)}))}{\sqrt[]{\sum^{n}_{i}(h(x_{i})-\overline{h(X)})^{2}\sum^{n}_{i}(h(y_{i}) -\overline{h(Y)})^{2}}}
\end{equation}

Where $X$ and $Y$ are the data sources we are trying to represent, $h(x_{i})$ and $h(y_{i})$ are the intermediate representations learned by the network for a given observation and $\overline{h(X)}$ and $\overline{h(Y)}$ are the mean intermediate representation for $X$ and $Y$, respectively.

\section{Proposed Model Architecture}

Given a parallel corpus our objective is training an encoder and decoder for each of the languages that are compatibles with the other components through a common intermediate representation generated by both encoders and understood by both decoders. For this, we propose a novel architecture and within it, we experiment with different distance measures and both discrete and continuous intermediate representations. 

\begin{figure}[h]
\centering
\includegraphics[scale=0.21]{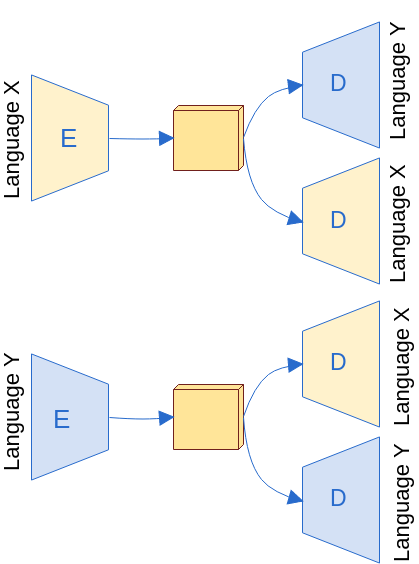}
\caption{Architecture example. Every module is compatible with the intermediate representation.}
\label{architecture}
\end{figure}

The architecture consists in an autoencoder for each of the languages to translate. Each network consists in a transformer network \cite{vaswani2017attention}. The advantage of using this model instead of other alternatives such as recurrent or convolutional encoders is that this model is based only on self attention and traditional attention over the whole representation created by the encoder. This allows us to easily employ the different components of the networks (encoder and decoder) as modules that during inference it can be used with other parts of the network without the need of previous step information as seen in Figure \ref{architecture}.

In order to achieve the desired universal language that can be used by all the modules of the system, all the components have to be optimized simultaneously. This is a difference to traditional NMT systems in which translation is only considered between the source and target language. In the proposed architecture, all languages are equally considered and both directions are generated during the training process. To achieve it, we design the following loss function:

\begin{equation}
\small{
Loss = L_{XX} + L_{YY} + L_{XY} + L_{YX} + d(h(X),h(Y))}
\end{equation}

Where $L_{XX}$ ($L_{YY}$) is the reconstruction loss of the autoencoder $X$ ($Y$) and $L_{XY}$ ($L_{YX}$) is the cross-entropy between the generated results from the encoder-decoder from language $X$ ($Y$) to language $Y$ ($X$) and the target reference in language $Y$ ($X$).

The final term of the loss is the measure of the distance between the two intermediate representations $h(X)$ and $h(Y)$ of both autoencoders. For this distance, we propose: 
\begin{enumerate}
\item\textit{Correlation distance} which measures how correlated are the intermediate representations of the autoencoders for each batch of the training process: 
\begin{equation}
\small
d(h(X),h(Y))  = 1 - c(h(X),h(Y))  
\end{equation}

\item \textit{Maximum distance} which measures the closeness of the intermediate representations as the maximum of the difference of the representation of a source and its target sentence:

\begin{equation}
\small
d(h(X),h(Y)) = max(|h(X)-h(Y)|)
\end{equation}
\end{enumerate}

For scaling to more languages, we will add language $Z$ training $L_{XZ}$ and $L_{ZX}$ (over a pretrained system). This implies that we do not require a multilingual parallel corpus on languages $X,Y,Z$ but only parallel corpus on languages $XY$ and $XZ$, for example.

\section{Evaluation of the Universal Language Representation}

Our main objective is creating an internal representation that can be understood by all the different modules trained in the system, where the modules are the encoders and decoders of all the languages involved in training. Similar representations may not lead to compatible encoder/decoders. Also different trainings can produce representations with different mean distances in the representations that can generate similar translation outputs.

In order to overcome those difficulties, we propose a new measure for the task. Given a parallel set of sentences in the languages in which the system has been trained, we can generate the encondings $E_{X}$ and $E_{Y}$. Both encodings, coming from parallel test, have the same number of vectors each of them of the same dimensionality. 

Our proposed measure consists in infering one of the decoders in the system ($X$ and $Y$) using $E_{X}$ and $E_{Y}$ as input. This generates two different outputs: an autoencoding output and a machine translation output. As we have parallel references for both languages we can measure the BLEU of each of the results against the reference to measure how the models perform. 

\begin{figure}[h!]
\centering
\includegraphics[scale=0.21]{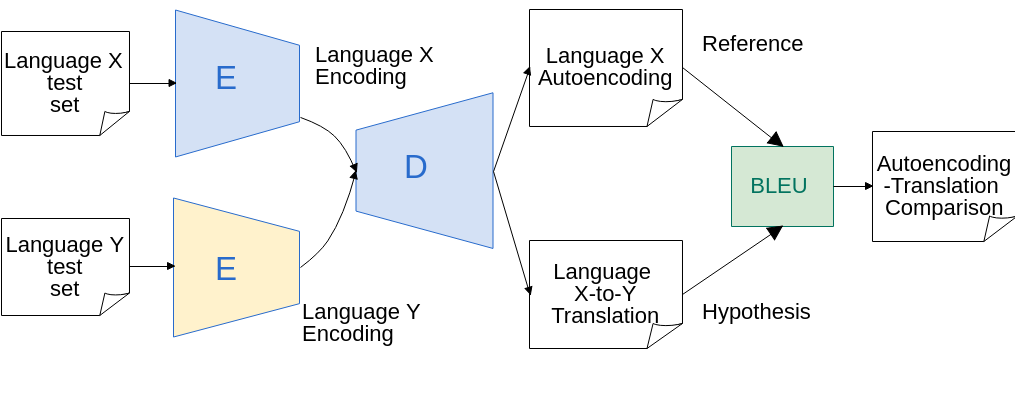}
\caption{Pipeline of the Interlingua BLEU measure.}
\label{measure}
\end{figure}

Additionally we can calculate a new BLEU score comparing the outputs of the autoencoding and the machine translation outputs. In the ideal case, encoders from two different languages have to produce the same representation for the same sentences. Therefore, the difference between the BLEU score obtained in the autoencoding output and the translation output shows how different are $E_{X}$ and $E_{Y}$ representations in terms of how the decoder is able to generate accurate results from them. Our measure consists in evaluating the BLEU score using the autoencoding output as reference and the machine translation output as hypothesis. Figure \ref{measure} shows the full pipeline of this procedure.

\section{Experimental framework}

For experiments, we use the Turkish-Engish parallel data from \textit{setimes2} \cite{tiedemann:2009} which is used in WMT 2017 \footnote{http://www.statmt.org/wmt17/}. The preprocess procedure consisted in limiting sentence length to 50 words, tokenization, truecasing using Moses tools \cite{koehn2007moses} and segmentation of Byte Pair Encoding (BPE) \cite{sennrich:2015}. As development and test set we used \textit{newsdev2016} and \textit{newstest2016}, respectively. 

All experiments were executed using the transformer model \cite{vaswani2017attention} with default parameters, 6 blocks of multihead attention of 8 heads each, 12,000 words vocabulary, latent representation size of 128 and fixed learning rate of 0.0001 using Adam \cite{kingma2014adam} as optimizer.

\section{Results}

\subsection{Translation quality}

Table \ref{results} shows the BLEU results in each translation direction from English-to-Turkish (EN-TR) and from Turkish-to-English (TR-EN). Results of different configurations of the proposed architecture (Univ) are compared to 
the baseline transformers (both non variational and variational, dvq) with the same hyperparameters of our architecture. 

\paragraph{Variational vs non-variational autoencoder.} The performance of the baseline transformer (non-variational) is almost competitive with the best system results from WMT 2017 \cite{mercedes:2017}. Note that we are comparing to the case of using only parallel data, without adding backtranslated monolingual data (which were 10.9 for EN-TR and 14.2 for TR-EN). When contrasting our proposed architectures, we see that the performance of non-variational autoencoders is higher than the variational autoencoder, with a difference of more than 1 \textit{BLEU} point in both directions.

\paragraph{Correlation vs Maximum distance loss.} In regard to the distance loss, the correlation distance clearly provides better translation results by approximately 1.5 \textit{BLEU} in both directions. Using correlation distance in fact shows similar performance compared to the one direction transformer baseline transformer. 

\begin{table}[!h]
\small
\centering
\caption{BLEU results for the different system alternatives, Transformer and different configurations of our architecture, Universal (Univ) with and without decomposed vector quatization (dvq), and correlation distance(corr) and maximum of difference(max)}
\label{results}
\begin{tabular}{l|l|l|}
\cline{2-3}
                                         & \textbf{EN-TR} & \textbf{TR-EN} \\ \hline
\multicolumn{1}{|l|}{Transformer}        & 8.32           & 12.03          \\ \hline
\multicolumn{1}{|l|}{Transformer dvq}    & 2.89           &  8.14          \\ \hline
\multicolumn{1}{|l|}{Univ + corr}       & 8.11           & 12.00          \\ \hline
\multicolumn{1}{|l|}{Univ + max}        & 6.19           & 10.38          \\ \hline
\multicolumn{1}{|l|}{Univ + dvq + corr} & 7.45           &  7.56          \\ \hline
\multicolumn{1}{|l|}{Univ + dvq + max}  & 2.40           &  5.24          \\ \hline
\end{tabular}
\vspace{-1em}
\end{table}

\subsection{Universal language representation quality}

We have also studied the difference in performance of decoders when presented with the universal representations of both encoders. This evaluation is performed in order to analyse if we can use an architecture of independent encoder/decoders in the context of MT. The model used for these results is the \textit{univ+corr}, which is the best performing model from Table \ref{results}. 

Table \ref{bleu-decoders} shows that the quality of the output of the decoder is quite better when the input comes from the encoder of the same language (autoencoder) than from another (MT). We also included the BLEU score between both autoencoder and translation outputs (A-T), which is the measure that we are proposing to evaluate the quality of our universal language. Low BLEUs in \textit{A-T} indicates that we are still far from being able to decode from the universal language.

\begin{table}[]
\small
\centering
\caption{Comparison of BLEU scores on the \textit{univ+corr} architecture when performing as autoencoder and MT. The third column is the BLEU between autoencoder and translation outputs}
\label{bleu-decoders}
\begin{tabular}{|l|l|l|l|}
\hline
\textbf{Decoder} & \textbf{Autoencoder} & \textbf{MT} & \textbf{A-T} \\ \hline
EN               &  63.32                    &  12.00                  &  11.90      \\ \hline
TR               &  59.33                    &   8.11                  &   6.02      \\ \hline
\end{tabular}
\end{table}
\normalsize	




\section{Visualization}

In this section, we can visualize the relation between the universal language representations.

Using the tool proposed at  \cite{ajenjo:2018} we employ UMAP \cite{mcinnes2018umap}. This technique computes a non linear dimensionality reduction of the data in order to be able to visualize the sentence representation space.

\begin{figure}[h]
\centering
\includegraphics[scale=0.25]{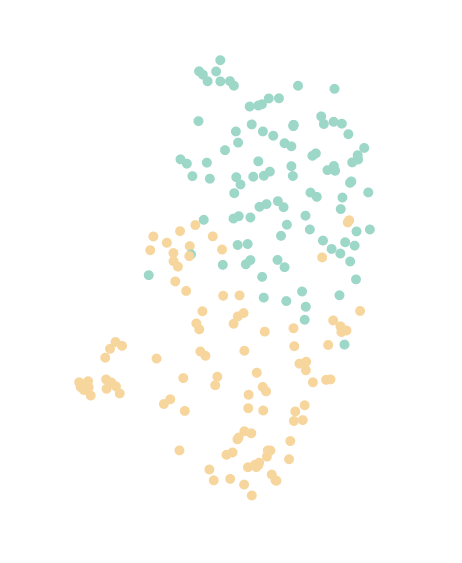}
\caption{Visualization of the sentence representation space using UMAP. Turkish in yellow, English in green.}
\label{visualization-space}
\end{figure}

Ideally if both encoders produce the same representation the visualization would show that both languages are no separable and that both dots for the same sentence would appear nearly in the same point in the plot. Figure \ref{visualization-space}  shows that languages appear to be located in different clusters. This difference in the representation can also explain the results in the previous section.  

\begin{figure}[h]
\centering
\includegraphics[scale=0.2]{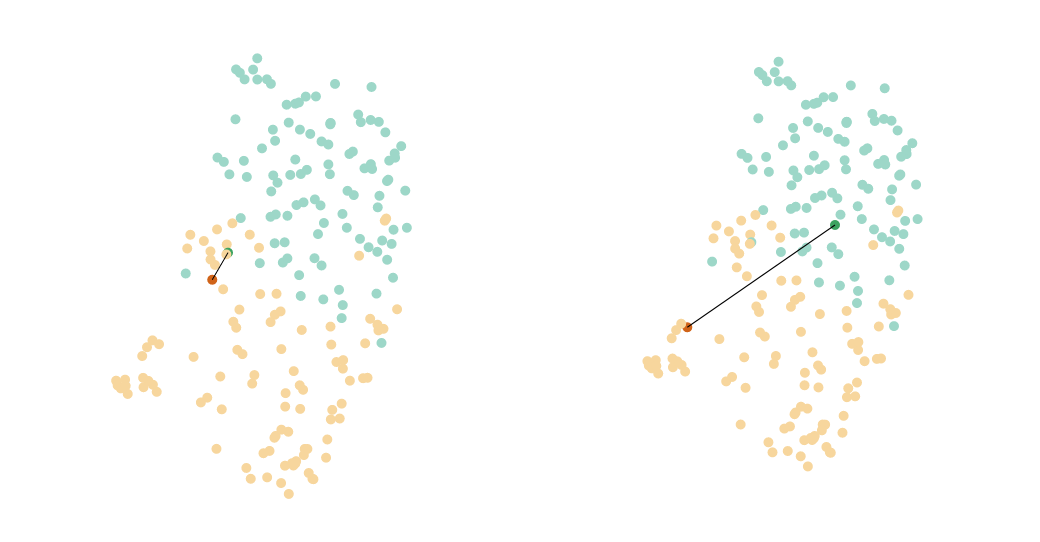}
\caption{Visualization of the sentence representation space using UMAP. Turkish in yellow, English in green.}
\label{visualization-examples}
\end{figure}

Focusing on the representation of individual sentences in the space it can be observed that the distance between parallel sentences is not constant as shown in Figure \ref{visualization-examples}.

\section{Conclusions}

We propose a novel translation architecture which includes a common intermediate representation as training objective. We explore both self-attentive variational and non-variational autoencoders to generate the universal language representation. The results show that in terms of quality, the proposed architecture is similar to the
baseline system but with the advantage of moving towards training an intermediate representation.

Innovatively, measuring generated outputs with the same decoding but using two language encodings shows that more work is still needed in order to produce the desired universal language representation for interlingual MT. 

As further work, we are exploring the use of the proposed architecture to better exploit limited resources. Additional encoders and decoders could be trained using only parallel corpus to one of the languages of a previously trained system, and then use the learned universal language representation to translate to and from all the languages already in the system.

\section*{Acknowledgements}

This work is supported in part by the Spanish Ministerio de Econom\'ia y Competitividad, the European Regional  Development  Fund  and  the  Agencia  Estatal  de  Investigaci\'on,  through  the  postdoctoral  senior grant Ram\'on y Cajal, the contract TEC2015-69266-P (MINECO/FEDER,EU) and the contract PCIN-2017-079 (AEI/MINECO).

\bibliography{emnlp2018,relwork}
\bibliographystyle{acl_natbib_nourl}


\end{document}